# Cognitive-mapping and contextual pyramid based Digital Elevation Model Registration and its effective storage using fractal based compression.

Suma Dawn
Jaypee Institute of
Information Technology,
Dept. of CS/IT,
A-10, Sec-62, Noida, India
suma.dawn@jiit.ac.in

Vikas Saxena
Jaypee Institute of
Information Technology,
Dept. of CS/IT,
A-10, Sec-62, Noida, India
vikas.saxena@jiit.ac.in

Bhudev Sharma
Jaypee Institute of Information Technology,
Dept. of Mathematics,
A-10, Sec-62, Noida, India
bhudev.sharma@jiit.ac.in

*Abstract:* Image Registration implies mapping images having varying orientation, multi-modal or multi-temporal images to map to one coordinate system. Digital Elevation models (DEM) are images having terrain information embedded into them. DEM-to-DEM registration incorporate registration of DEMs having different orientation, may have been mapped at different times, or may have been processed using different resolutions. Though very important only a handful of methods for DEM registration exist, most of which are for DEM-to-topographical map or DEM-to-Remote Sensed Image registration.

Using cognitive mapping concepts for DEM registration, has evolved from this basic idea of using the mapping between the space to objects and defining their relationships to form the basic landmarks that need to be marked, stored and manipulated in and about the environment or other candidate environments, namely, in our case, the DEMs. The progressive two-level encapsulation of methods of geo-spatial cognition includes landmark knowledge and layout knowledge and can be useful for DEM registration. Space-based approach, that emphasizes on explicit extent of the environment under consideration, and object-based approach, that emphasizes on the relationships between objects in the local environment being the two paradigms of cognitive mapping can be methodically integrated in this three-architecture for DEM registration. Initially, P-model based segmentation is performed followed by landmark formation for contextual mapping that uses contextual pyramid formation. Apart from landmarks being used for registration key-point finding, Euclidean distance based deformation calculation has been used for transformation and change detection.

Initially, P-model based segmentation is performed followed by landmark formation for contextual mapping that uses contextual pyramid formation. Landmarks have been categorized to belong to either being flat-plain areas without much variation in the land heights; peaks that can be found when there is gradual increase in height as compared to the flat areas; valleys, marked with gradual decrease in the height seen in DEM; and finally, ripple areas with very shallow crests and nadirs. For the final storage of co-registered DEMs, fractal based compression has been found to give good results in terms of space and computation requirements.

In this paper, an attempt has been made to implement DEM-DEM registration based on human spatial cognition method of recollection. This method may further be extended for DEM-to-topographic map and DEM-to-remote sensed image registration. Experimental results further cement the fact that DEM registration may be effectively done using the proposed method.

## 1. INTRODUCTION

DEM (Digital Elevation Model) data files consist of only the elevation or height values of the terrain, covering a specified area in a discreet grid-like 3-D space of the particular surface in consideration. DEMs can be useful for extracting and visualization of terrain parameters, cartographic map generation and updation, modeling water flow or mass movement amongst others [26], [27], [32].

DEM registration, in general, is a method of overlaying two or more DEMs, or a map. DEM registration is important as it allows for seamless integration of DEMs the same place which may be represented in different orientation, at different times, or may have been processed using different resolutions. Most methods are used for image registration in general that have been emulated in [48 - 52].There are only a handful of methods for DEM registration most of which are for DEM-to-topographical map or DEM-to-Remote Sensed Image registration. DEM-to-DEM registration is elusive because of the difficulties of finding feature points or control points required for matching and evaluation and error analysis [**30**].

Human spatial cognition is an interdisciplinary research area in cognitive science and encompasses some of these theoretic and technologic methods of acquiring, managing, visualization, communication and service of geospatial knowledge etc. Route establishment and way-finding typically requires planning and the ability to stay oriented while moving [**17**], mostly incorporated with the usage of 'landmarks' or 'key-reference-points'.

Using cognitive mapping concept for DEM registration has evolved from this basic idea of using the mapping between the space to objects and defining their relationships to form the basic landmarks that need to be acquired and manipulated in and about the environment and route-finding in adjacent or other candidate environment, in our case DEMs. Landmarks or 'key-reference-points' serve as anchor points to entertain local environments in one's cognitive maps and such a method of equation of existence of certain landmarks in one map to other will serve as a complete route finding and for the purpose of registration of these entities. Further on, if learning-based systems were to be used, it would ensure that a repeated encounter with prominent landmarks, so identified, would make the system much faster for registration of DEMs containing such related objects.

The basic 2-tier geo-spatial cognition entails landmark knowledge and layout, considering the content it expresses [**17**]. This progressive encapsulation method can be useful for DEM registration where-in the landmarks or key-feature-points and their subordinate-points may be used for depicting the three-tier architecture for making the registration process much robust as well as well-defined. Two paradigms of cognitive mapping exist: space-based approach, emphasizing on explicit extent of the environment under consideration, and object-based approach, emphasizing on the relationships between objects in the local environment [20]. Such paradigms can be methodically integrated in this three-architecture for DEM registration.

In this paper, an attempt has been made to implement DEM-DEM registration based human spatial cognition method of way-finding and recollection. This may further be extended for DEM-to-topographic map and DEM-to-remote sensed image registration. Also route finding using the landmarks have been attempted.

Landmarks may be seen to belong to four categories in DEMs: flat -plain areas without much variation in the land heights; peak formation with gradual increase in height as compared to the flat areas; valley formation with gradual decrease in the height seen in DEM; ripples with very shallow crests and nadirs. The landmark type formation works on the assumption of certain threshold values used, the values of which may be changed if the number of landmarks found is not adequate for a given DEM.

DEM registration between two DEMs, say DEMA and DEMB has been implemented in a three fold algorithm. Firstly designating and storing the landmarks. These have been formed by using pyramid manner of calculation.

Secondly, matching the various landmarks based on their types. Finally, after orientation determination of the candidate DEM with respect to the reference DEM, registration of both the DEMs is performed. If landmarks cannot be formed, contours have been used for anchor point detection.

Route-finding or way-finding between two points in a given single DEM or in multiple DEMs ( wherein say point A lies in DEMA and point B lies in DEMB) have been implemented by first registering the DEM if points lie in different DEMs and then retrieving the landmarks from the knowledgebase followed by path-finding and route depiction

The remainder of the paper is organized in the following manner: Section II contains the various related terms used in the work. Section III presents our proposed methodology for digital elevation model registration using cognitive method by landmark detection and inexact graph matching and further on, route finding between given two points while Section IV presents the experimental results for both the registration process as well as route-finding method. In Section V we have summarized the literature presented and also outlined our future work.

## 2. PRESENT STATE OF THE ART

DEM registration is important as it allows for seamless integration of DEMs of the same locality which may have been represented in different orientation, or may have been processed using different resolutions. Most DEM registration methods include DEM-to-topographical map registration or DEM-to-Remote Sensed Images registration and have evolved from methods used for generation and mosaicing of DEMs [33], [35]. There are only a handful of methods for DEM-to-DEM registration. DEM-to-DEM registration is elusive because of the difficulties of finding feature points or control points required for matching and evaluation and error analysis of the techniques so used.

Digital elevation models present the bare earth height model. Among the enumerable applications, the generic applications of DEM include - extracting terrain parameters, cartographic map generation and updation, modeling water flow or mass movement (say, avalanches and landslides), creation of relief maps, rendering of 3D visualizations including flight planning, creation of physical models, rectification of aerial photography or satellite imagery, terrain analyses in geomorphology and physical geography, Geographic Information Systems (GIS), engineering and infrastructure design, Global Positioning Systems (GPS), line-of-sight analysis, precision farming and forestry, Intelligent Transportation Systems (ITS), Advanced Driver Assistance Systems (ADAS). DEM's are also utilized in support of the pre-planning and lay-out of corridor surveys, seismic line locations, construction activities, etc [**26**],[**27**], [**32**].DEMs may be illustrated as either depiction of height values only or a 3D view of these mappings.

A comparison of the various methods involving elevation models is presented in Table 5. It also includes comparison to our presented work. These papers include comparison from [26] - [28], [30], [31], [33] - [41].

The methods studied above for DEM-with-DEM fusion or DEM-with-other images, suffer from drawbacks like time and complexity intensive, lack in error matrix evaluation, applicability to only a few chosen images of particular resolution or types amongst other. Also, none prove to be suitable for different resolution DEMs. DEM-to-DEM registration is elusive because of the difficulties of finding feature points or control points required for matching and evaluation and error analysis [30].The study performed for this paper proposes a novel method for DEM-to-DEM registration. Using cognitive mapping concept for DEM registration has evolved from this basic idea of using the mapping between the space to objects and defining their relationships to form the basic landmarks that need to be acquired and manipulated in and about the environment and route-finding in adjacent or other candidate environment, in our case DEMs. Landmarks or 'key-reference-points' serve as anchor points to entertain local environments in one's cognitive maps and such a method of equation of existence of certain landmarks in one map to other will serve as a complete route finding and for the purpose of registration of these entities. Further on, if learning-based systems were to be used, it would ensure that a repeated encounter with prominent landmarks would make the system much faster for registration of DEMs containing such related objects.

In this paper, an attempt has been made to implement DEM-DEM registration based human spatial cognition method of way-finding and recollection. This may further be extended for DEM-to-topographic map and DEM-to-remote sensed image registration.

## 3. PROPOSED METHODOLOGY

DEM-to-DEM registration has been attempted in this work. Cognitive map concept has been used for DEM registration due to its inherent relationship to marking or forming landmarks for identification in the active environment. Two paradigms of cognitive mapping: space-based approach, emphasizing on explicit extent of the environment under consideration, and object-based approach, emphasizing on the relationships between objects in the local environment, exist [20]. Orientation and distance finding may not be in exact metric terms, rather in semantic terms of number of hops of smaller landmarks, etc [12].

The notion of key-reference-point can be successfully used for landmark forming and storing in the DEMs considered which form the basis for their registration. If supervised-learning system could be adopted, repeated detection of the same landmarks would make the system more robust and faster for registration.

Spatial cognition perception includes attributes like locations, separation and connection, size, directions, distances, shapes, patterns and movements. It allows cognitive agents to act and interact in space intelligently and to communicate about spatial environments in meaningful ways. Cognitive mapping and geo-spatial visualization use the cognitive behavior that has been vastly studied [1] – [6], [15] - [18], [21]. The cognition of geo-space can be divided as a progressive process into landmark knowledge, route knowledge and layout knowledge - three levels, considering the content it expresses [17]. Cognitive mapping of such system basically works in three stages or key assertions. Firstly, understanding or knowing the work space through contrast and similarity, i.e. what is the work space or area? Secondly, understanding of the problem or application requirements, i.e. what is to be done and defining the problem space. And thirdly, for seeking solutions of the defined space by defining hierarchical constructs. Some of these constructs may be depicted to be super-ordinate to others, whereas some may be deemed to be of the same category, like those forming an organizational design.

For our working, we have categorized the landmarks as per their prominent geographical feature. They are (i) flat or plain areas – areas without much variation in their height; (ii) peak – area with gradual increase in the pyramid formation; (iii) valley – areas with gradual decrease in the height found through pyramid calculations; (iv) and ripple areas – like sand dunes / small hillocks with very shallow crests and nadirs. These groupings have been done with the aid of threshold values. The landmark type formation works on the assumption of certain threshold values used, the values of which may be changed if the number of landmarks found is not adequate for a given DEM.

DEM registration between two DEMs, say $DEM_{ref}$ and $DEM_{cand}$, reference and candidate DEMs respectively, have been implemented in a three fold algorithm. Firstly designating and storing the landmarks. These have been formed by using pyramid manner of calculation. Secondly, matching the various landmarks based on their types. Finally, after orientation determination of the candidate DEM with respect to the reference DEM, registration of both the DEMs is performed. If landmarks cannot be formed, contours have been used for anchor point detection.

---

**Algorithm** DEM_registration($DEM_{ref}$, $DEM_{cand}$, $DEM_{registered}$)

Input: Two DEMs namely, reference DEM, and candidate DEM - $DEM_{ref}$, $DEM_{cand}$.
Output: registered DEM data file - $DEM_{registered}$.

Step 1: Preprocessing & p-model based segmentation
Step 2 Landmark detection based on the contextual pyramid formation. This step uses cognitive-pyramid formation for extraction of contextual information. This classification may be extended using fuzzy classification techniques as well.
Step 3: Classify landmarks, form landmark-based graphs and insert into landmark knowledge base. The above 3 steps are followed only if any or both the DEMs to be registered are not already present in the knowledgebase.
Step 4: Find the maximum matched sub-graph
Step 5: find orientation and deformations between the maximum matched sub-graphs.
Step 6: match landmark graphs based on sub-graph matching
Step 7: Register the candidate and reference DEMs.

---

Certain preprocessing steps to landmark detection have been performed. P-model based segmentation has been shown to give better results as compared to simple morphology based segmentation. These segments are then fed to the fuzzy c-means based contextual pyramid forming engine to perform contextual pyramid-based landmark formation and detection.

P-algorithm is a modification of watershed and then waterfall models that are used for segmentation. The difference is in the fact that in P algorithm, all the initial contours of the given signal are compared, at each hierarchical level, to the hierarchical image. Also, in P algorithm, not only the contours inside potential maxima-islands but also, the contours embedded in catchment basins are considered. This algorithm is used for its pyramid and hierarchical structure kind of property as the signal structure could be classified and used as per their two external categories of one corresponding to a classical mosaic structure where the lower levels of hierarchy may be embedded in to their corresponding higher ones and secondly, the signal structure made of maxima-islands.

The landmark type formation works on the assumption of certain threshold values used. These values may be changed if the number of landmarks found is not adequate for a given DEM.

Contour lines and graph node have been used as landmarks. These have been formed based on contextual pixel classification. Contextual classifiers and contextual re-classifiers have been used to form the landmarks and the contour points. Such calculations need to be done for all the candidates and the reference DEMs used for registration purpose.

The Graph Isomorphism problem (GI) tests whether two given graphs are isomorphic or not. Various details have been dealt by authors in [42] - [44], [46] - [47]. If there are a large number of graphs, subgraph query [44], may be shown as one of the most fundamental procedures in managing graphs and may be used for querying for patterns from large networks. Since in the proposed algorithm, the graphs of each landmark category may or may not contain the same number of nodes in corresponding reference and candidate DEMs, they would be considered as cases for the inexact graph matching. For such class of problems, exact isomorphism, may or may not be found. However, the aim would be to find the best possible matching [46] with maximum number of nodes of the candidate DEM matching to those of reference DEM. This class of problem would be applied for the cases wherein if the set of vertices for candidate DEM is defined as VC and the set of vertices for the reference DEM is said to be VR, then the matching would be performed for cases |VM| < | VR| and separately for | VR| <|VM|, giving rise to the need for solution through sub-graph matching. For such cases, dummy nodes have been considered to complete the number of nodes, for matching purpose, where ever lesser number of nodes was found [46]. This gives rise to the position-oriented graphs.

The position relates to the orientation of the matched sub-graph and is used for finding global transformation.

After registration, for effective storage, fractal based compression has been used.

## 4. EXPERIMENTAL RESULTS & DISCUSSION

For our experiments, we have used DEMs of different places. Each set has been so chosen that there are some common areas so that registration would result in the display the common as well the disparity between the DEMs considered. For computation of the moving concentric window for pyramid-based computation, increasing the concentric window size beyond the computed windows do not yield any new information along with the analysis becoming unnecessarily computationally complex, the window size was restricted to 10 x 10. Every DEM is considered to be having at least 3 major landmarks of the same class. If such major landmarks are not being able to be formed, the initial threshold levels are readjusted.

Random sets of DEMs were chosen for experimentation. Their resolution and related data are shown in Table I. The proposed technique was implemented and tested on 5 sets of DEMs each set containing 5 reference and 5 candidate DEMs, making a total of 25 DEMs.

In the sets used for experimentation, we have also considered the % of common area and the number of landmarks as shown in table II. As seen from the below shown table, if the common area is approximately less than 60-65 % of common area, the number of landmarks that could be used for the further tasks of registration is insufficient and hence are not used. For better sets of results, the actual % of common area used for registration is in the range above 70%.

From the considered data set, one from each class of set that have been used and their registered data, are shown below. One set of reference, candidate, and their corresponding registered DEMs have been shown. In Figure 1, the third column shows the Output data of registration. These sets belong to the range of 70% to 95% common area.

The below shown table, Table III gives the counts of the various landmark types Peak, valley, flat-plain and ripple found after being processed for the DEM sets shown in the above fig 1.

The registration process must eventually be evaluated against a method to satisfy the similarity measure between the candidate DEM and registered DEM. Performance of the registration method, so proposed, has been evaluated by the most simplest and easily calculable maximum correlation coefficient. The whole candidate DEM may or may not be registered to the reference DEM, ie., exact and complete matching and registration may not be possible. Therefore, to accommodate the inexact matching and registration process's evaluation, the mutual information of the common area found in the registered DEM, and candidate DEM and the reference DEM's have been compared.

The method presented in this paper has been tested using Correlation Coefficient similarity metric that gives an easy indication of its merit. As shown from the result set of table IV, the technique is found to give good results for registration.

The proposed method also has the merits as compared to other methods by having more comprehensible and logical anchor-point detection method, ease with post-disaster registration of DEMs. Also, DEM specification is not a constraint for the proposed technique. And multi-modal and multi-temporal DEMs can be easily registered using the said approach. Though having high time and computation complexity, the model is much robust as compared to other feature finding techniques.

Robustness of the proposed algorithm is checked by introducing Gaussian noise of random nature of range -10 to +10 and zero mean and then its performance are evaluated, as shown in Table V. The graph shown in Figure 2 confirms the proposed theory that that the proposed algorithm performs better as compared to the three commonly used methods for registration namely, transform domain-based registration, KLD-based registration, and iterative closest point-based registration.

The below shown table, table VI shows the comparison of various similarity measures used for evaluation of performance showing for various amounts of common areas for various sets of DEMs used for registration.

Table VII shows the comparison of various compression methods used for compressing the registered file. The properties of compression ratio and peak signal-to-noise ratio between before compression and after decompression have been used for comparison. The graph for the same is shown in fig 3.

Comparison with some of the other methods proposed by various other authors have also been done. This is emulated in table VIII.

The major problems arise due to the border data which cannot be fully used for finding the landmarks as these are based on contextual pyramid formation which require larger bases. As the number of iterations grows, the base of the contextual pyramid also grows. Hence, the initial time complexity for the complete registration process is high. However, as shown from the comparisons made above once the DEM is included in the landmark knowledge base, its re-computation time for registration reduces drastically.

## 5. CONCLUSION & FUTURE RESEARCH

If a cluster of similar landmark nodes are found, these can be grouped together by virtue of their proximity, and such cluster may be considered as a major landmark, and the centroid of this cluster would then be constructed and marked as a single landmark rather than a group of similar-landmark-nodes. The vicinity may be decided based on the proximity to the nearest-similar landmarks. If there are more than 3 same class landmarks within a radii threshold distance, they, together, would be grouped together to form a major landmark.

The basic steps of geo-spatial cognition entails three levels of knowledge - landmark knowledge, route knowledge and layout knowledge - considering the content it expresses. This progressive encapsulation method can be useful for DEM registration where-in the landmarks or key-feature-points and their subordinate-points may be used for depicting the three-tier architecture for making the registration process much robust as well as well-defined. In this paper, an attempt has been made to implement DEM-DEM registration based human spatial cognition method of recollection. Further on, if learning-based systems would ensure that a repeated encounter with prominent landmarks would make the system much faster for registration of DEMs containing such related objects.

For our experimentation, landmarks have been categorized as flat-plain areas, peaks, valleys, and ripple areas based on their inherent characteristics of elevations. The landmark type formation works on the assumption of certain threshold values used, the values of which may be changed if the number of landmarks found is not adequate for a given DEM. The first stage consists of initial finding and classification of the various landmarks found in each of the DEMs. This forms the landmark knowledge-base. For registration, the reference and candidate DEMs are then compared based on the various landmarks found. The matching process is further sustained by the characteristics of the landmarks. Partial and non-exact graph matching is used for the above requirement.

The experiment has involved five sets of reference and candidate DEMs to be registered, each set having 5 test data sets i.e., five reference and five candidate DEMs, making a total of 25 DEMs. Random sets of DEMs were chosen for experimentation. As indicated by the normalized mutual information values, the algorithms performance is found to be adequately good in terms of the final output. Experimentation was also performed for various percentages of common area for evaluating the robustness of the proposed algorithm. Robustness measure based on CC and MI values of some of the common methods after adding Gaussian Noise to check the robustness of the proposed algorithm. Comparison with some of the other methods proposed by various other authors was also discussed.

TABLES & FIGURES

| SETS | RESOLUTION | | |
|---|---|---|---|
| | Units – meters, Arc sec | x | y |
| Set 1 | ~30 m, 1 arc-sec | 512 | 512 |
| Set 2 | ~90 m, 3 arc-sec | 900 | 900 |
| Set 3 | ~90 m, 3 arc-sec | 1200 | 1200 |
| Set 4 | ~90 m, 3 arc-sec | 1500 | 1500 |
| Set 5 | ~900 m, 30 arc-sec | 3000 | 3000 |

Table I. Description of the DEMs used in Experiment

| Amount of common area | Resolution | No. of various Landmarks found for matching | | | |
|---|---|---|---|---|---|
| | | Peaks | Valleys | Flat | Ripple |
| 10 % | 1500 x 1500 | 2 | 5 | 1 | 26 |
| 20% | 900 x 900 | 1 | 0 | 3 | 13 |
| 30% | 3000 x 3000 | 6 | 19 | 15 | 41 |
| 40% | 1200 x 1200 | 12 | 11 | 9 | 26 |
| 45% | 512 x 512 | 3 | 9 | 18 | 27 |
| 50% | 900 x 900 | 12 | 9 | 36 | 45 |
| 55% | 3000 x 3000 | 44 | 17 | 64 | 51 |
| 60% | 1500 x 1500 | 91 | 42 | 131 | 124 |
| 65% | 1200 x 1200 | 8 | 12 | 16 | 27 |
| 70% | 512 x 512 | 6 | 11 | 18 | 27 |
| 75% | 1500 x 1500 | 102 | 49 | 186 | 174 |
| 80% | 512 x 512 | 11 | 14 | 19 | 29 |
| 84% | 900 x 900 | 15 | 19 | 121 | 65 |
| 88% | 3000 x 3000 | 106 | 68 | 319 | 203 |
| 92% | 1500 x 1500 | 147 | 52 | 224 | 217 |
| 96% | 512 x 512 | 15 | 19 | 21 | 35 |
| 100% | 1200 x 1200 | 208 | 135 | 97 | 113 |

Table II. Number of various landmarks found for matching depending on the % of common area for a set of DEMs used.

| SETS | | Average No. of Landmarks found for each category | | | | Landmark considered for registration | No. of points used for orientation determination |
|---|---|---|---|---|---|---|---|
| | | Peak | Valley | Flat | Ripple | | |
| Set 1 | Reference DEM | 49 | 13 | 9 | 23 | Peak | 35 |
| | Candidate DEM | 42 | 8 | 10 | 16 | | |
| Set 2 | Reference DEM | 14 | 23 | 107 | 189 | Ripple | 143 |
| | Candidate DEM | 19 | 21 | 115 | 195 | | |
| Set 3 | Reference DEM | 229 | 157 | 98 | 128 | Valley | 103 |
| | Candidate DEM | 146 | 173 | 105 | 113 | | |
| Set 4 | Reference DEM | 158 | 56 | 361 | 264 | Flat | 198 |
| | Candidate DEM | 172 | 51 | 295 | 261 | | |
| Set 5 | Reference DEM | 107 | 69 | 413 | 319 | Flat | 346 |
| | Candidate DEM | 104 | 82 | 521 | 298 | | |

Table III. Count of the various Landmark types – Peak, valley, flat-plain and ripple. Also shown is the output of the no of maximum matches found after performing sub-graph matching used for the various landmark types of each set of DEM data.

| Size of File | Proposed method | | transform domain-based registration | | Direct KLD-based registration | | Iterative closest point based registration | |
|---|---|---|---|---|---|---|---|---|
| | CC | MI | CC | MI | CC | MI | CC | MI |
| Set 1 | 0.93 | 1.43 | 0.77 | 0.87 | 0.83 | 1.27 | 0.75 | 0.85 |
| Set 2 | 0.80 | 1.39 | 0.72 | 0.81 | 0.75 | 1.38 | 0.72 | 0.83 |
| Set 3 | 0.83 | 1.58 | 0.78 | 0.91 | 0.83 | 1.49 | 0.79 | 0.89 |
| Set 4 | 0.90 | 1.25 | 0.74 | 0.81 | 0.87 | 1.01 | 0.79 | 0.77 |
| Set 5 | 0.93 | 1.41 | 0.70 | 0.89 | 0.87 | 1.32 | 0.68 | 0.80 |

Table IV. Comparative evaluation based on average CC and MI values of some of the common methods for an average of 90% common area coverage belonging to DEMs of various Sets.

| Set of DEM | Proposed method | transform domain-based registration | Direct KLD-based registration | Iterative closest point based registration |
|---|---|---|---|---|
| Set 1 | 1.148 | 0.61 | 0.73 | 0.55 |
| Set 2 | 1.038 | 0.65 | 1.08 | 0.59 |
| Set 3 | 1.281 | 0.71 | 0.79 | 0.89 |
| Set 4 | 1.160 | 0.76 | 0.91 | 0.69 |
| Set 5 | 1.103 | 0.69 | 1.02 | 0.69 |

Table V. Robustness measure based on Mutual Information measure of some of the common methods after adding Gaussian Noise of the range of +- 10 and zero mean and then performance evaluation.

| DEM set used | % of common area in reference & candidate DEM | Performance Evaluation | | |
|---|---|---|---|---|
| | | CC | MI | KLD |
| | 50% | 0.1152 | 0.442 | 0.23 |
| | 70% | 0.2807 | 0.78 | 0.55 |
| | 80% | 0.5285 | 1.01 | 0.87 |
| Set 1 | 90% | 0.8707 | 1.4321 | 1.1231 |
| | 50% | 0.2001 | 0.399 | 0.31 |
| | 70% | 0.5002 | 0.704 | 0.63 |
| | 80% | 0.7998 | 1.1011 | 0.81 |
| Set 2 | 90% | 0.8815 | 1.3945 | 1.2755 |
| | 50% | 0.2663 | 0.61 | 0.499 |
| | 70% | 0.7983 | 0.87 | 0.68 |
| | 80% | 0.8001 | 1.14 | 0.899 |
| Set 3 | 90% | 0.8707 | 1.5801 | 1.4811 |
| | 50% | 0.1109 | 0.52 | 0.441 |
| | 70% | 0.1532 | 0.8 | 0.69 |
| | 80% | 0.5602 | 1.001 | 0.89 |
| Set 4 | 90% | 0.8914 | 1.2522 | 1.221 |
| | 50% | 0.1532 | 0.7011 | 0.51 |
| | 70% | 0.2312 | 0.88 | 0.72 |
| | 80% | 0.7668 | 1.159 | 0.931 |
| Set 5 | 90% | 0.8889 | 1.4131 | 1.3021 |

Table VI. Table showing comparison of similarity metric values for the common area of the reference and the registered DEMs using correlation coefficient, mutual information and Kulback-Lieblier distance measures.

| Set of DEM | Average Compression ratio | | | Average Peak Signal-to-noise Ratio | | |
|---|---|---|---|---|---|---|
| | Fractal-based compression | Wavelet Transform coding | JPEG coding | Fractal-based compression | Wavelet Transform coding | JPEG coding |
| Set 1 | 2.1 | 1.88 | 2.1 | 78.25 | 65 | 71 |
| Set 2 | 2.56 | 2.5 | 2.23 | 79 | 65 | 73.56 |
| Set 3 | 3.09 | 2.85 | 2.79 | 83.54 | 69.12 | 75.3 |
| Set 4 | 2.81 | 2.6 | 2.63 | 80.91 | 68.84 | 74 |
| Set 5 | 2.75 | 2.6 | 2.54 | 79.88 | 68 | 74 |

Table VII. Table showing the comparison of various compression methods used for compressing the registered file. Criteria used for evaluation are average compression ratio and average peak signal-to-noise ratio ( PSNR)

| Methods by other authors | Work proposed by the authors | Feature Detection and Extraction methods used | Searching and Feature Matching methods used | Image Types used for experimentation | Similarity measure / Analysis method |
|---|---|---|---|---|---|
| Sefercik [26] | DEM generation from topographic maps | Geodesic instruments were used with the topographic maps to form contour-based extraction of heights. Stereo pairs were used for height information generation. | Shifting and superimposed data done using DEMSIFT program. | OrbView-3 space image. LIDAR and InSAR images were also used | DEMANAL program used for checking SRTM X-band height model |
| Trisakti and Carolita [27] | DEM generation | Ground Control Points (GCP) based generation | Not mentioned particularly | ASTER Stereo Data based on IKONOS image and SRTM | Not mentioned particularly |
| Li and Bethel [28] | Alignment and Registration of DEMs | Ground Control Point (GCP) based detection - 2.5D polynomial transformation for DEM registration. | GCPs used for matching | Interferometric SAR DEMs | Minimum RMSE based alignment |
| Yong and Huayi [29] | DEM generation | Morphological gradient based extraction – point cloud method | Filtering algorithm proposed | LIDAR data | Type I, Type II and Type III ( total error) assessment. |
| Maire & Datcu [31] | Integration of DEM and EO data for 3D rendering | Region extraction based on segmentation and dynamic generation of object-oriented image description to reflect geometry and topology. – Byesian approach, Gaussian Markov Random Fields | Interactive selection of regions and classification among a set of user-thematic. | X-SRTM DEM data, InSAR DEM. | Tree Structure Modeling |
| Futamura, Takaku, Suzuki, Iijima, Tadono, Matsuoka, Shimada, Igarashi, and Shibasaki [33] | High resolution DEM generation | GCP based orientation and orthophoto correction | Coarse-to-Fine Processing and Area-based stereo matching. | JERS-1/OPS data, PRISM data | DEM-histogram based analysis |
| Ferretti, Monti-Guarnieri, Prati, and Rocca [34] | DEM reconstruction from multiple images | Permanent Scatterers based reconstruction | Not mentioned particularly | SAR DEM, SPOT DEM. | Not mentioned particularly |
| Allievi, Ferretti, Prati, Ratti, and Rocca [35] | DEM reconstruction | Multi-interferogram and phase unwrapping approach | Multi-Baseline PU algorithm | ERS Tandem Pairs | Layover and comparison with prior topographic data |
| Ferretti, Prati, and Rocca [36] | DEM reconstruction | Wavelet domain – weighted average based reconstruction | Not mentioned particularly | InSAR DEM | Variance comparison |
| Saadi, Aboud, and Watanabe [37] | DEM, ETM+, Geologic, and Magnetic Data Integration | Shaded Relief Maps, Slop Maps, Traverse Profiles, | pseudogravity transformation | Landsat ETM+ images, Geophysical data, aeromagnetic data | Abrupt change in the magnetic anomaly and its analysis |
| Takcuchi [39] | Usage of DEM and slant range information for Image registration | Inverse mapping of foreshortening simulation method | Not mentioned particularly | SAR & TM data | Study of approximate errors by affine transformation. |
| Schultz, Riseman, Stolle, and Woo [40] | DEM fusion and generation of 3D terrain model | Self-consistency distribution based methodology | Ground control point based matching | DEMs | Self-consistency threshold |
| Lahoche, and Herlin [41] | Fusion of various image data for high resolution map generation of land surface temperature | Individual temporal profile estimation | Classification of land cover | Landsat TM, NOAA/AVHRR and DEM Data | Statistical validation by mean and confidence interval. |
| Proposed Methodology | DEM registration | Landmark based knowledge-base formation | Graph formation and inexact graph matching | DEMs | Correlation coefficient, Mutual Information, & Kullback-Lieblier Distance. |

Table VIII: Table showing the comparison of methods proposed by various authors for DEM or other image registration

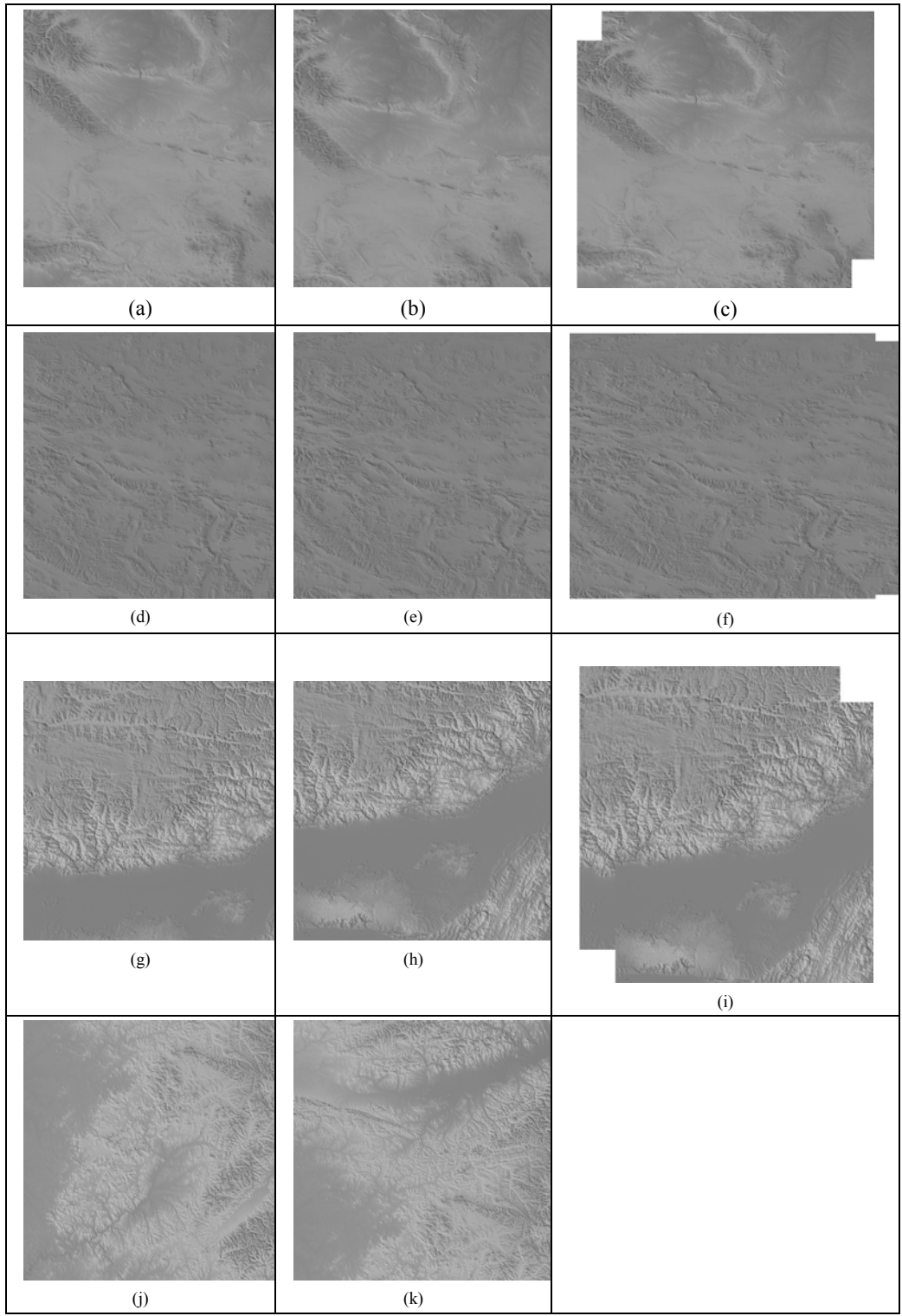

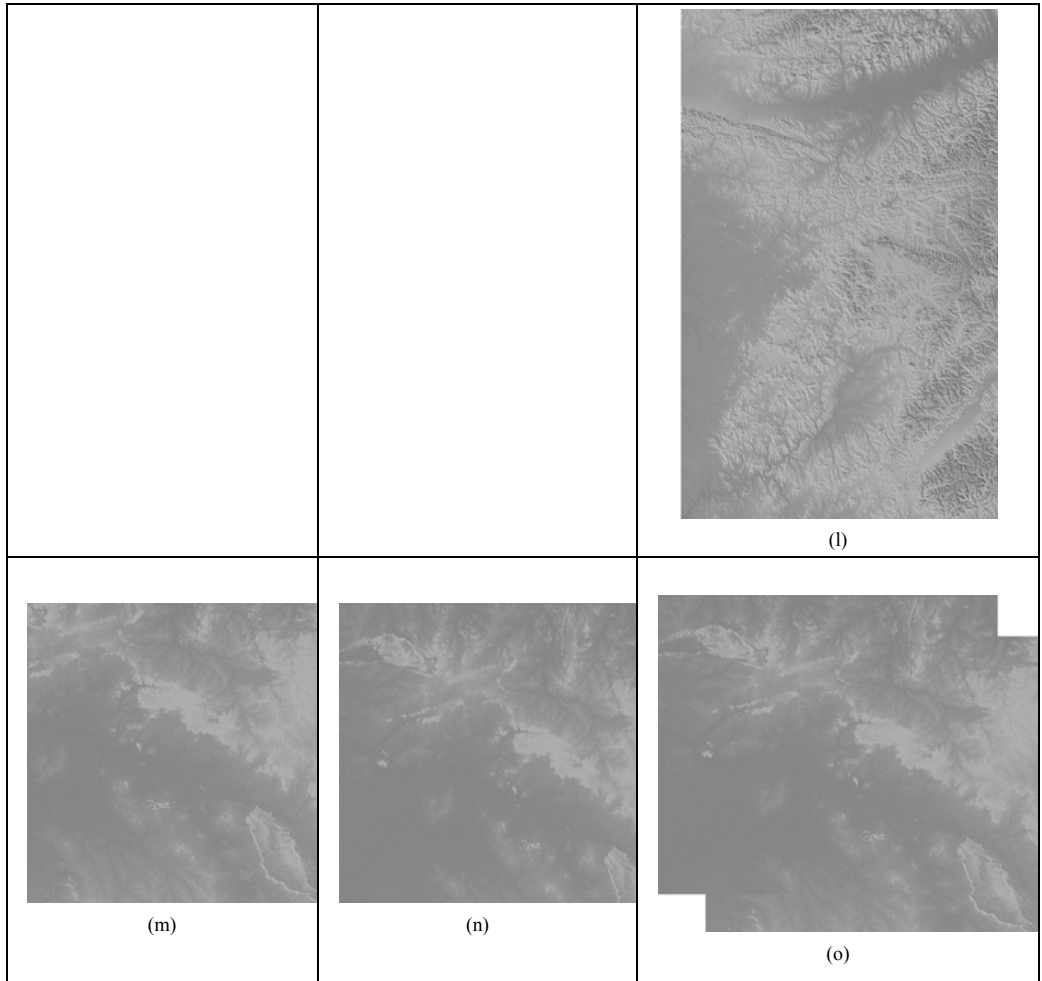

Figure 1. Left to Right, (a) to (j) – reference, candidate and registered DEMs of sets 1 to 5 respectively. Candidate and Reference DEM data curtsey - http://data.geocomm.com/dem/

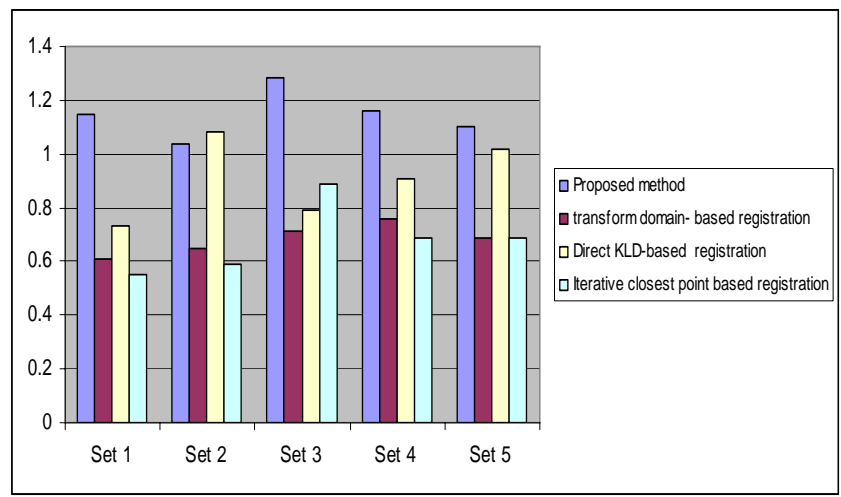

Figure 2. Graph plot showing the comparison for robustness measure to Gaussian noise of proposed method with those of transform domain-based, direct KLD-based and Iterative Closest point-based registration based on Mutual Information (MI) measure.

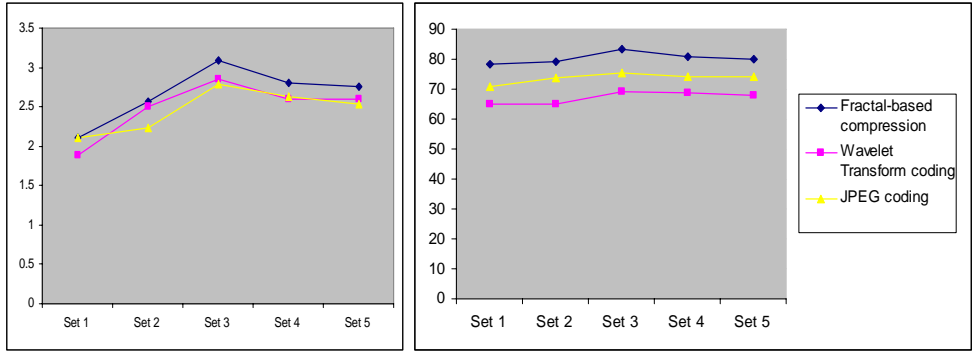

Fig. 3. Graphs showing the trends of Fractal-based compression to those of wavelet transform coding and JPEG coding based compression with respect to 3(a) Average Compression Ratio, and 3(b) Average PSNR